\documentclass[10pt,twocolumn,twoside]{IEEEtran}
\newcommand{\journalname}{Journal Name}
\usepackage{tmi}
\usepackage{cite}
\usepackage{url}
\usepackage[hidelinks]{hyperref}
\hypersetup{
    colorlinks=true,        
    linkcolor=blue,
    citecolor=blue,
    urlcolor=blue,
}
\usepackage{amsmath,amssymb,amsfonts}
\usepackage{algorithmic}
\usepackage{graphicx}
\usepackage{textcomp}
\usepackage{amssymb} 
\usepackage{multirow}
\usepackage{booktabs}
\usepackage{makecell}
\usepackage{bm}
\usepackage{amsmath} 
\usepackage{siunitx}
\usepackage{subcaption}

\def\BibTeX{{\rm B\kern-.05em{\sc i\kern-.025em b}\kern-.08em
    T\kern-.1667em\lower.7ex\hbox{E}\kern-.125emX}}
\begin{document}
\title{BrainCast: A Spatio-Temporal Forecasting Model for Whole-Brain fMRI Time Series Prediction}
\author{Yunlong Gao, Jinbo Yang, Li Xiao, Haiye Huo, Yang Ji, Hao Wang, Aiying Zhang, and Yu-Ping Wang 
\thanks{This work was supported in part by the National Natural Science Foundation of China under Grants 62202442, 12261059, and 62401535, in part by the Natural Science Foundation of Jiangxi Province under Grant 20224BAB211001, in part by the Institute of Artificial Intelligence of Hefei Comprehensive National Science Center under Grants 23YGXT004 and Y1022502, in part by the NIH under Grants R01GM109068, R01MH104680, R01MH107354, R01EB006841, P20GM103472, and U19AG055373, and in part by the NSF under Grant 1539067.}
\thanks{Y. Gao and J. Yang are with the Institute of Advanced Technology, University of Science and Technology of China, Hefei 230052, China.}
\thanks{L. Xiao and Y. Ji are with the MoE Key Laboratory of Brain-Inspired Intelligence Perception and Cognition, University of Science and Technology of China, Hefei 230052, China (e-mail: xiaoli11@ustc.edu.cn).}
\thanks{H. Huo is with the School of Mathematics and Computer Sciences, Nanchang University, Nanchang 330031, China.}
\thanks{H. Wang is with the MoE Key Laboratory of Brain-Inspired Intelligence Perception and Cognition, University of Science and Technology of China, Hefei 230052, and also with the Anhui Province Key Laboratory of Biomedical Imaging and Intelligent Processing, HFCNS Institute of Artificial Intelligence, Hefei 230088, China.}
\thanks{A. Zhang is with the School of Data Science, University of Virginia, Charlottesville, VA 22903, USA.}
\thanks{Y.-P. Wang is with the Department of Biomedical Engineering, Tulane University, New Orleans, LA 70118, USA.}
}

\maketitle

\begin{abstract}
Functional magnetic resonance imaging (fMRI) enables noninvasive investigation of brain function, while short clin-\\ical scan durations, arising from human and non-human factors, usually lead to reduced data quality and limited statistical power for neuroimaging research. In this paper, we propose BrainCast, a novel spatio-temporal forecasting framework specifically tailored for whole-brain fMRI time series forecasting, to extend informative fMRI time series without additional data acquisition. It for-\\mulates fMRI time series forecasting as a multivariate time series prediction task and jointly models temporal dynamics within reg-\\ions of interest (ROIs) and spatial interactions across ROIs. Spec-\\ifically, BrainCast integrates a Spatial Interaction Awareness mo-\\dule to characterize inter-ROI dependencies via embedding every ROI time series as a token, a Temporal Feature Refinement mod-\\ule to capture intrinsic neural dynamics within each ROI by enh-\\ancing both low- and high-energy temporal components of fMRI time series at the ROI level, and a Spatio-temporal Pattern Align-\\ment module to combine spatial and temporal representations for producing informative whole-brain features. Experimental results\\ on resting-state and task fMRI datasets from the Human Connec-\\tome Project demonstrate the superiority of BrainCast over state-of-the-art time series forecasting baselines. Moreover, fMRI time series extended by BrainCast improve downstream cognitive abil-\\ity prediction, highlighting the clinical and neuroscientific impact brought by whole-brain fMRI time series forecasting in scenarios with restricted scan durations.
\end{abstract}

\begin{IEEEkeywords}
Attention,
fMRI,
spatio-temporal modeling,
time series forecasting
\end{IEEEkeywords}

\section{Introduction}\label{sec:introduction}
Functional magnetic resonance imaging (fMRI), by measuring spontaneous fluctuations in blood oxygen level-dependent (BOLD) signals, has emerged as a preferred tool for functional imaging in noninvasive neuroscience studies \cite{Fox2007Spontaneous,Buckner2008The}. It enables the characterization of brain activity and large-scale functional connectivity, providing critical insights into both normal cogn-\\ition and neurological disorders \cite{Logothetis2001Neurophysiological,Logothetis2008What}. Despite its broad app-\\lications, a persistent challenge in clinical settings is that fMRI scan durations are often short, due to participant-related factors (e.g., reduced tolerance for long-duration scans in psychiatric, pregnant, or younger and elder participants) and practical cons-\\straints (e.g., scan cost or protocol limitations) \cite{b3,b4}. Short-duration scans typically result in lower-quality fMRI data, cha-\\racterized by reduced signal-to-noise ratio, compromised test–\\retest reliability, and diminished sensitivity in downstream ana-\\lyses \cite{b5,b6,b7}. Although extending fMRI scan duration can alle-\\viate these issues, longer acquisitions will increase participant burden and also the susceptibility to motion artifacts, and thus are not generally feasible in routine clinical practice \cite{b8}. Rec-\\ent evidence further suggests that longer scans increase statis-\\tical power and cost-efficiency in whole-brain association studies \cite{Ooi2025Longer}. These observations collectively motivate the development of methodology based approaches that can effectively ex-\\tend fMRI time series without additional data acquisition.

Beyond data augmentation, the ability to predict future brain activity from historical fMRI data offers a complementary and potentially powerful means of understanding neural dynamics. On the other hand, accurate fMRI time series forecasting may facilitate the study of intrinsic brain process, enhance the mod-\\eling of interregional interactions, and enable applications such as cognitive assessment, disease phenotyping, and brain–com-\\puter interface. From a signal analysis perspective, fMRI time series exhibit strong temporal correlation and structured fluct-\\uations, indicating that future fMRI data is partially predictable from previous observations \cite{Sobczak2021Predicting,Hjelm2018Spatio}. However, to the best of our knowledge, few methods have been specifically developed for fMRI time series forecasting.

Recent advances in deep learning have led to substantial pr-\\ogresses in generic time series forecasting, particularly in capt-\\uring complicated temporal dependencies. For instance, recurr-\\ent neural network (RNN) and long short-term memory (LST-\\M) models can model short-term temporal dependencies \cite{b9,b10}, while Transformer-based architectures leverage attention mechanisms to learn long-range temporal dependencies \cite{Vaswani2017Attention,Zhou2021Informer,Zhang2023Crossformer}. These models have demonstrated notable success in time series prediction across domains such as traffic flow and power consumption. Nevertheless, they are not tailored for fMRI time series forecasting, in which fMRI time series are characterized by considerable noise, nonstationary dynamics, and rich inter-regional dependencies. For such data, forecasting performance relies not only on temporal modeling within individual regions but also on considering interactions across brain regions.

In \cite{b17}, BrainLM was built on a masked autoencoder frame-\\work for large-scale self-supervised pretraining, treating future fMRI time series prediction as an auxiliary task without explicitly modeling inter-region interactions. Generative Adversarial\\ Networks (GANs) have been investigated for fMRI time series forecasting, but were restricted to low-frequency components from a single region within the default mode network (DMN), instead of whole-brain data \cite{b38}. Although more recent studies have attempted to account for spatio-temporal dependencies in fMRI data  by graph neural network (GNN)- and Transformer-based models \cite{b14,b15,b16}, their focus has primarily been on adv-\\anced representation learning for brain functional connectivity analyses, rather than explicit fMRI time series forecasting. As\\ a result, whole-brain fMRI time series forecasting remains ins-\\sufficiently explored, highlighting the need for joint coordination of temporal dynamics and spatial interactions.

\begin{figure*}[!t]
\centerline{\includegraphics[width=0.92\textwidth]{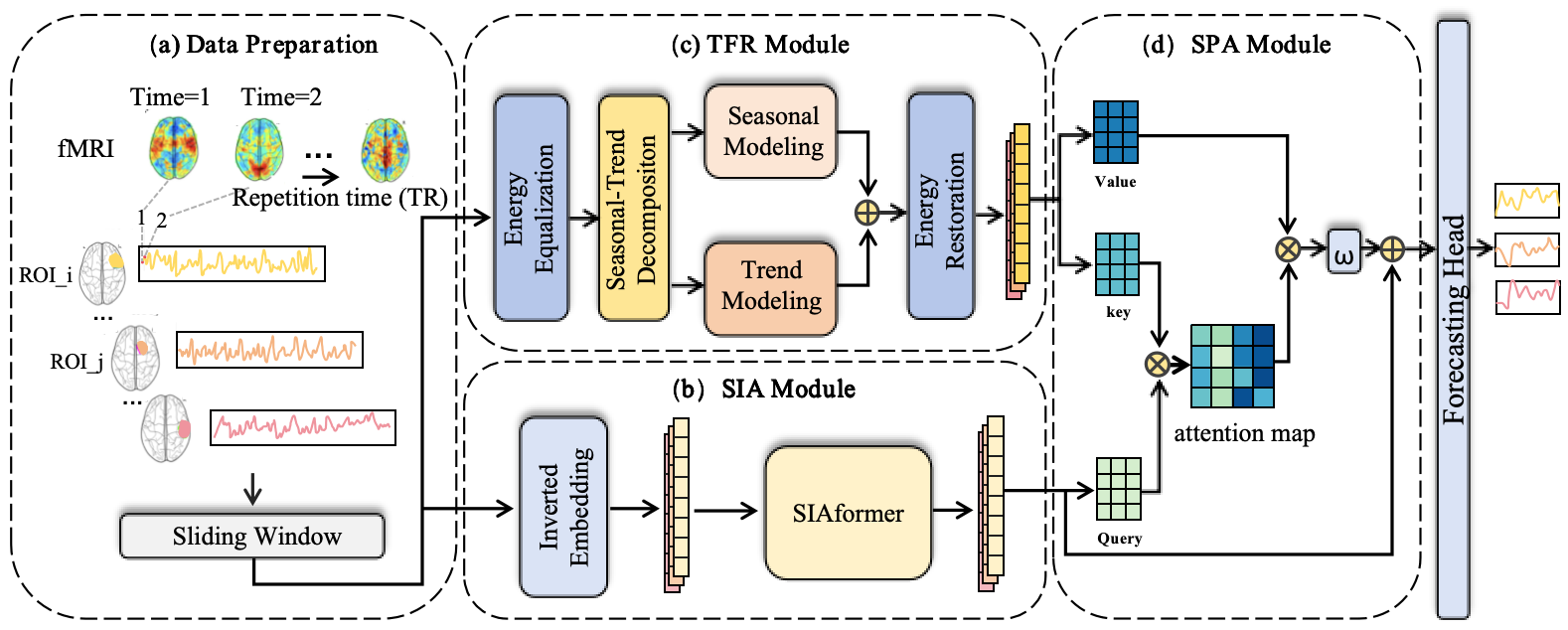}}
\caption{The overall framework of BrainCast. (a) fMRI data preprocessing; (b) The Spatial Interaction Awareness (SIA) module for learning between-ROI spatial interactions; (c) The Temporal Feature Refinement (TFR) module for extracting temporal dy-\\namics within every ROI; and (d) The Spatio-temporal Pattern Alignment (SPA) module for aligning spatio-temporal represent-\\ations across ROIs and time points.}
\label{fig1}
\end{figure*}

In this paper, we address this gap via proposing BrainCast, a spatio-temporal forecasting framework specifically designed for whole-brain fMRI time series forecasting. As illustrated in\\ Fig. \ref{fig1}, BrainCast utilizes a sliding-window partition strategy to formulate forecasting tasks and integrates spatial and temporal information in a unified architecture that forecasts whole-brain fMRI time series over a future horizon, given prior fMRI time series from a predefined look-back. BrainCast consists of three main modules: i) a Spatial Interaction Awareness (SIA) modu-\\le used to explicitly characterize inter–region-of-interest (inter-ROI) interactions by embedding each ROI (variate) time series as a token; ii) a Temporal Feature Refinement (TFR) module introduced to enhance both low- and high-energy temporal co-\\mponents for capturing intrinsic neural dynamics within every ROI; and iii) a Spatio-temporal Pattern Alignment (SPA) mod-\\ule designed to align spatial and temporal representations from\\ SIA and TFR, respectively, for producing informative whole-brain features, followed by a prediction head that forecasts the target fMRI time series. In addition, extensive experiments on\\ resting-state and task fMRI datasets from the Human Connect-\\ome Project (HCP) \cite{b18} demonstrate the superiority of BrainCast over state-of-the-art time series forecasting baselines. Fur-\\thermore, extending fMRI time series using BrainCast results in more accurate cognitive ability prediction, underscoring the broader potential of fMRI time series forecasting for neurosci-\\ence research and clinical applications.

The contributions of this paper can be summarized below.

\begin{itemize}
  \item This paper is among the first to examine fMRI time series\\ forecasting as a multivariate time series prediction probl-\\em and to jointly model inter-ROI spatial interactions and intra-ROI temporal dynamics in a unified architecture. 
  \item We propose BrainCast tailored for whole-brain fMRI time\\ series forecasting, which incorporates SIA, TFR, and SPA modules to account for inter-ROI dependencies, nonstati-\\onary temporal dynamics, and heterogeneous spatio–tem-\\poral representations, respectively, for improved forecasting accuracy.
  \item Experimental results on large-scale fMRI datasets demonstrate that BrainCast outperforms several competitive time\\ series forecasting models across multiple evaluation metrics. In addition, we show that fMRI time series extended\\ using BrainCast lead to improved cognitive ability predic-\\tion, providing empirical evidence that whole-brain fMRI forecasting can enhance brain–cognition analyses, partic-\\ularly under limited fMRI scan durations.
\end{itemize}

The remainder of this paper is organized as follows. Section \ref{sec2} reviews related studies on time series forecasting and brain\\ fMRI signal analysis. Section \ref{sec3} presents the proposed BrainCast framework in detail. Section \ref{sec4} reports experimental res-\\ults on fMRI datasets from the HCP, followed by discussions\\ on visualization, hyperparameter influence, and applications in cognitive ability prediction in Section \ref{sec5}. Section \ref{sec6} concludes this paper.

\section{Related Work}\label{sec2}
\subsection{Time Series Forecasting}
Deep learning models have demonstrated great potential in time series forecasting. For example, RNN-based models, such as RNNs and LSTMs\cite{b9,b10}, have been presented through the extraction of temporal information from historical data for predicting future data, yet they are ordinarily limited in captur-\\ing long-term dependencies within time series. To address this\\ issue, recent studies have begun to explore alternative architectures, e.g., Temporal Convolutional Network and Transformer\cite{b19,Vaswani2017Attention}. Particularly, the self-attention mechanism in Transformer enables effective modeling of long-range dependencies. Early Transformer-based models \cite{b20,Zhou2022Fedformer} treated all variables at the same time point as one single token, which often led to semantic entanglement among variables and suboptimal perf-\\ormance. To mitigate this, PatchTST \cite{b22} presented a channel-\\independent approach that processed every variable separately. While this process improves clarity in variable representations, it incurs increased training time and fails to model interactions between variables. In \cite{Liu2023itransformer}, iTransformer was proposed to over-\\come these limitations by regarding each complete time series as a token, which allows the model to effectively capture inter-variable relationships. In parallel, GNN-based models have pr-\\oven well suited in exploiting coupled temporal and spatial de-\\pendencies \cite{b24,b25}. Collectively, these advances have led to significant progress in time series forecasting and/or analyses. However, the majority of these above models were introduced for domains such as traffic flow and electricity usage, and are not specifically tailored to fMRI data. By contrast, fMRI time\\ series exhibit different traits, e.g., substantial noise, nonstationary dynamics, and complicated inter-ROI interactions, which pose unique challenges for time series forecasting that are not adequately accounted for by existing forecasting models.

\subsection{Brain fMRI Data Analysis}
In recent years, significant advances have been made in the application of fMRI analysis techniques. Convolutional Neural\\ Networks (CNNs) were among the earliest deep learning models applied to the analysis of diverse neurological disorders and\\ brain connectivity patterns, such as BrainNetCNN and Convol-\\utional Auto-Encoder \cite{b26,b27,b28,b29}, and have demonstrated enco-\\uraging performance. In addition, GNNs have displayed strong potential in brain connectivity network analysis owing to their powerful capability to capture local structural information bet-\\ween ROIs \cite{b30,b31,b32}. For instance, BrainGNN adopted graph convolutional layers to enable biomarker prediction at both the\\ population and individual levels \cite{b33}. Dynamic GNNs further exploited the dynamics of brain connectivity to consider time-varying interactions between ROIs \cite{b34}. More recently, Trans-\\former-based models have attracted increasing attention owing to their superior ability to learn long-range dependencies \cite{b35,b36}, and have been successfully applied in fMRI data analysis \cite{b37}. Although the above mentioned models have significantly advanced fMRI data analysis, most existing studies remain fo-\\cus on characterizing observed fMRI data, with limited attention to modeling and forecasting future whole-brain fMRI time series. Remarkably, BrainLM leveraged a masked autoencoder architecture to pre-train a self-supervised model on large-scale fMRI datasets \cite{b17}, where future fMRI time series forecasting served as an auxiliary task and inter-ROI interactions were not\\ explicitly emphasized. In \cite{b38}, a GAN-based model was devel-\\oped for forecasting future fMRI time series; however, the for-\\ecasting was restricted to low-frequency components of fMRI time series from one single ROI within the DMN, rather than whole-brain fMRI time series.

\section{Methods}\label{sec3}

\subsection{Problem Definition}
\begin{figure}[!h]
\centerline{\includegraphics[width=0.85\columnwidth]{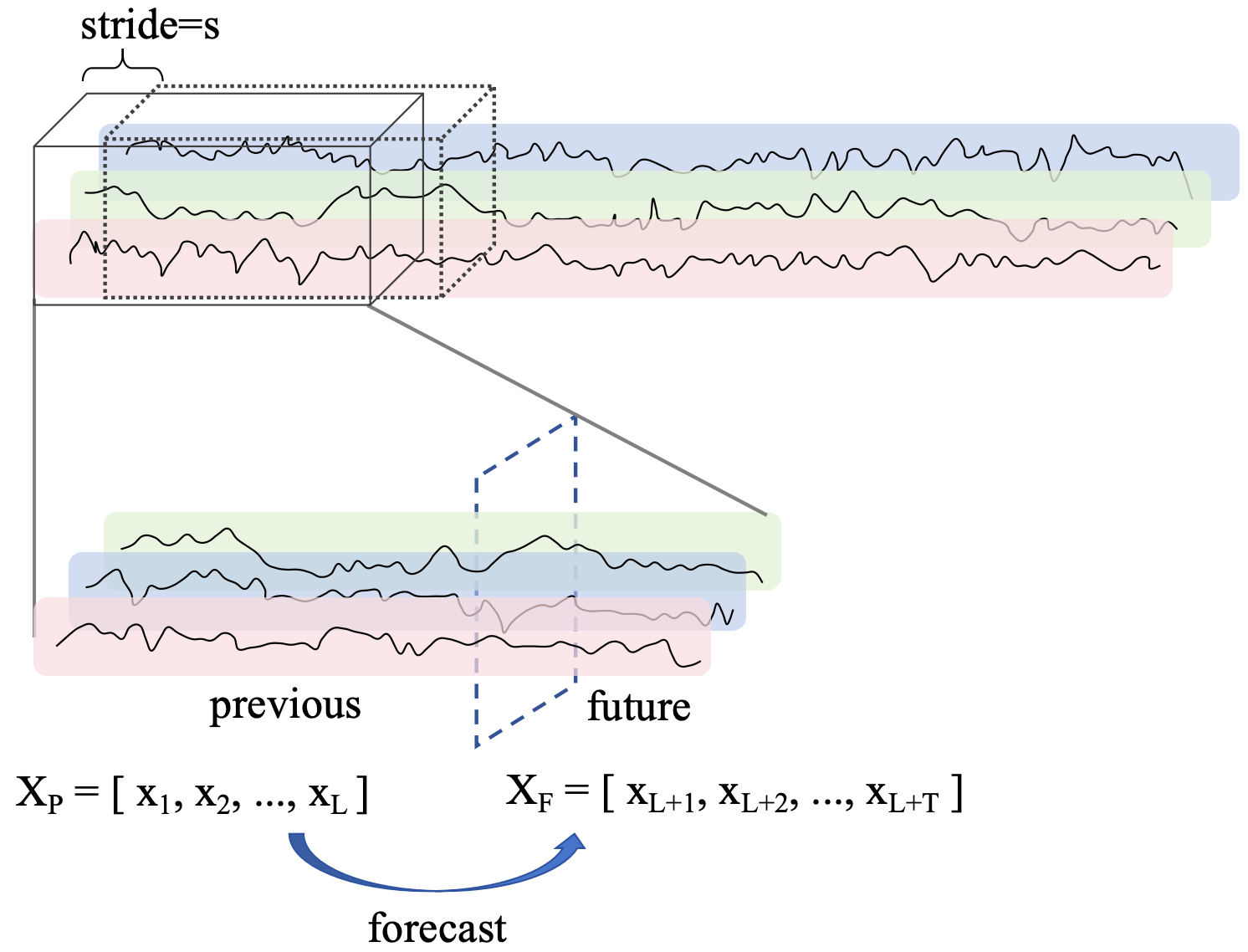}}
\caption{A sliding window partition technique on the fMRI data to generate sequential samples.}
\label{slidwin}
\end{figure}

Given fMRI data of a subject, it is a multivariate time series with $N$ variables, in which the $N$ variables correspond to the pre-defined $N$ ROIs, and the $i$-th univariate time series is the observed fMRI responses of the $i$-th ROI over a period of run duration for $1\leq i\leq N$. We apply a sliding window partition technique on the fMRI data to generate sequential samples, as illustrated in Fig. \ref{slidwin}, where the window size is set to $L+T$, the stride is set to $s$, and the fMRI time series over every sliding window is referred to as a sample. For each sample, let the first $L$-point fMRI time series $\bm{X}_P = [\bm{x}_1, \bm{x}_2, \cdots, \bm{x}_L]\in\mathbb{R}^{N\times L}$ be the previous data, and the following $T$-point fMRI time series $\bm{X}_F = [\bm{x}_{L+1}, \bm{x}_{L+2}, \cdots, \bm{x}_{L+T}]\in\mathbb{R}^{N\times T}$ be the future data, where $\bm{x}_i\in\mathbb{R}^N$ represents the observed fMRI responses of all ROIs at the $i$-th time point. The fMRI time series forecasting task we study in this paper is to establish a mapping from the previous data $\bm{X}_{P}$ to the future data $\bm{X}_{F}$. 

\begin{figure*}[!t]
\centerline{\includegraphics[width=0.7\textwidth]{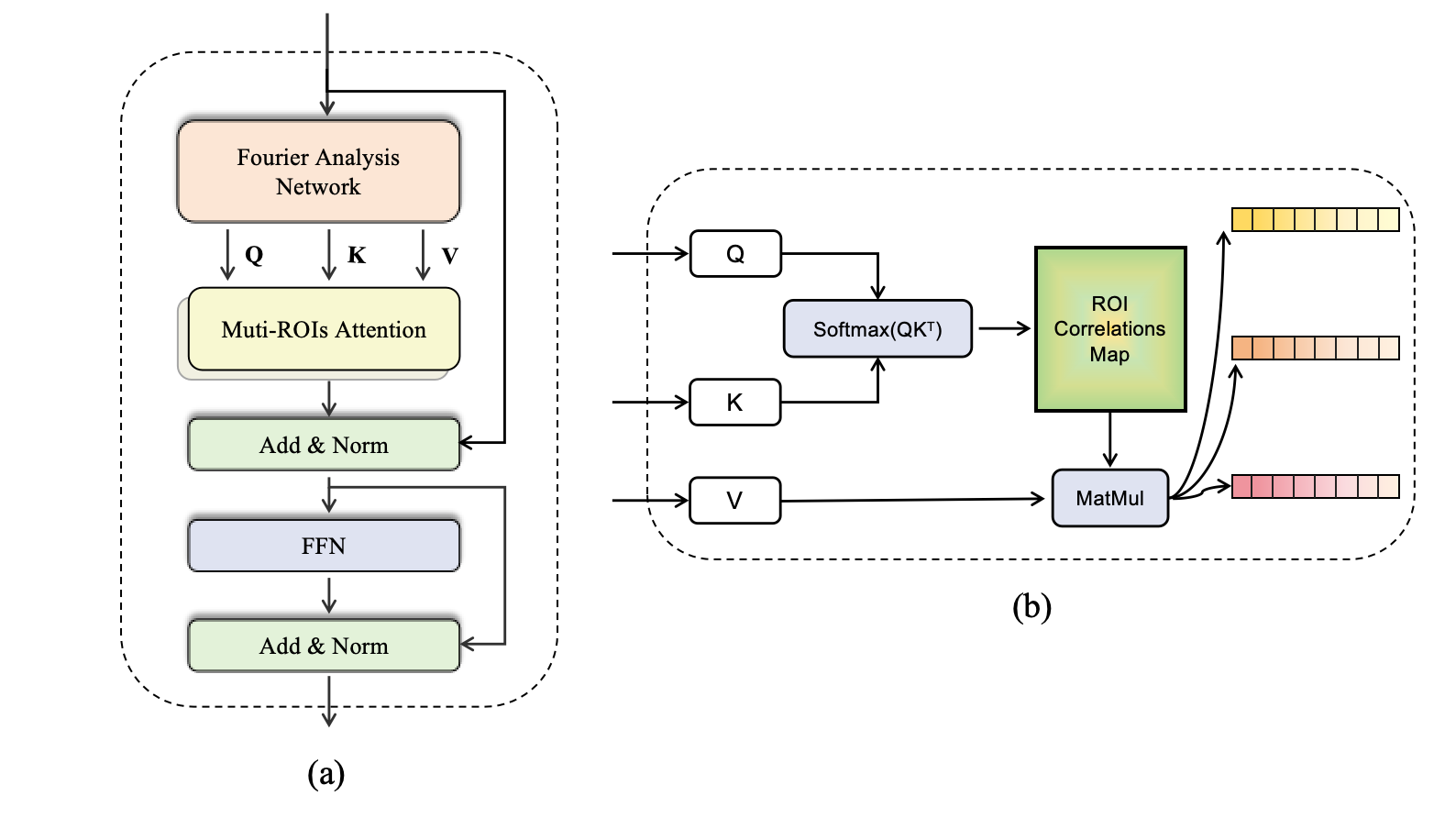}}
\caption{(a) Illustration of SIAformer, in which self-attention (b) is applied to learning spatial interactions between ROIs.}
\label{siaformer}
\end{figure*}

To do so, we propose BrainCast, a spatio-temporal forecasting model for brain fMRI time series prediction. As shown in Fig. \ref{fig1},  BrainCast includes three important components, i.e., a Spatial Interaction Awareness (SIA) module formulated to un-\\cover direct spatial interactions between distinct ROIs; a Tem-\\poral Feature Refinement (TFR) module tailored to capture the intrinsic temporal dynamics within every ROI; and a Spatio-\\temporal Pattern Alignment (SPA) module leveraged to align heterogeneous spatial and temporal features for robust whole-brain representations. Benefiting from this architecture, BrainCast enables accurate, reliable prediction of multi-region fMRI time series. The details of the three modules are stated below.


\subsection{Spatial Interaction Awareness}
The conventional Transformer-based time series forecasters, such as Informer \cite{Zhou2021Informer} and Autoformer\cite{Wu2021Autoformer}, typically formed a (temporal) token by multiple variables of the same time point and performed the generative formulation of forecasting tasks. \\ However, they may not be suitable for multivariate time series forecasting, particularly for the aforementioned fMRI time ser-\\ies forecasting task, due to the following reasons. First, fMRI\\ data is unavoidably coupled with noise (which may be physio-\\logical, instrumental or motion-related) during image acquisi-\\tion. This implies that a certain number of strong-noise tokens are mixed with all the tokens, and regarding all the tokens equ-\\ally will potentially hinder the learning of true neural represen-\\tations. Second, Transformer using the temporal tokens results in meaningless attention maps, and fails to capture multivariate correlations that are significant for characterizing between-ROI interactions in fMRI based brain network analysis.

Therefore, inspired by iTransformer \cite{Liu2023itransformer}, we design an SIA module (see Fig. \ref{fig1}(b)) to capture spatial dependencies between ROIs, i.e., multivariate correlations, where each univariate time series is embedded as a (variate) token, and the obtained vari-\\ate tokens are implemented by our customized SIAformer for multivariate correlating and are individually input into a shared FFN for learning nonlinear representations. More specifically, we first embed each univariate time series in $\bm{X}_P$ as a variate token of dimension $D$ by a shared MLP, i.e.,
\begin{equation}\label{embedd}
\bm{X}_E\in\mathbb{R}^{N\times D}:=\text{Embedding}(\bm{X}_P).
\end{equation} 
The obtained variate tokens $\bm{X}_E$ aggregate global information of time series over time, which can be better used by attention mechanisms in SIAformer for multivariate correlating below.

We subsequently use a stack of $G$ SIAformer layers on $\bm{X}_E$, each composed of Fourier Analysis Network (FAN) \cite{b47}, self-\\ attention, and FFN, as shown in Fig. \ref{siaformer}. More concretely, based\\ on $\bm{X}_E$, FAN is employed to derive query, key, and value matr-\\ices ($\bm{Q}, \bm{K}, \bm{V}\in\mathbb{R}^{N\times D}$), i.e.,
\begin{equation}\label{qkv}
\left(\bm{Q},\bm{K},\bm{V}\right)=\text{FAN}(\bm{X}_E),
\end{equation}
respectively. The reason we use FAN here is that FAN can eff-\\ectively extract implicit periodic patterns and regularities from the variate tokens through learnable spectral filters\cite{Chen2015BOLD}. Based on the attention mechanism, we have the attention map matrix $\bm{A}\in\mathbb{R}^{N\times N}$ and token embedding matrix $\bm{H}\in\mathbb{R}^{N\times D}$ as
\begin{equation}\label{attention}
\bm{A} = \text{Softmax}\left( \bm{Q}\bm{K}^\top/\sqrt{D} \right) \;\text{and} \;\bm{H} = \bm{A}\bm{V},
\end{equation}
respectively. Of note, since each variate token corresponds to a specific ROI, the resulting attention map matrix $\bm{A}$ can reflect the dependencies  between ROIs. A shared FFN is applied to\\ the embedding of each variate token in $\bm{H}$ to further integrate relational information. In addition, as a basic operation widely used to enhance the convergence and training stability of deep neural networks, layer normalization (LN) is also performed.\\ Finally, a comprehensive representation $\bm{H}^{\text{spat}}\in\mathbb{R}^{N\times D}$ that incorporates spatial interactions between ROIs is given by
\begin{equation}\label{sss}
\bm{H}^{\text{spat}} = \text{LN}\left(\text{FFN}(\bm{H})+\text{LN}(\bm{H}+\bm{X}_E)\right),
\end{equation}
which follows a gated or residual transformation.

\subsection{Temporal Feature Refinement}
To capture the temporal characteristics of fMRI time series, we present a TFR module using frequency domain techniques. Most frequency domain based time series forecasting methods\\ tend to focus on high-energy components (low-frequency part)\\ while ignoring low-energy components (high-frequency part), leading to sub-optimal prediction performance. Moreover, low-energy components of fMRI time series encode stable patterns of spontaneous neural activities \cite{Fox2007Spontaneous}, making them essential for fMRI time series forecasting. Therefore, we use Amplifier \cite{Fei2025Amplifier} as our TFR module (see Fig. \ref{fig1}(c)), because Amplifier can learn both low- and high-energy components without bias, as briefly stated in what follows.

We first equalize low- and high-energy components through spectrum flipping. Let $\bm{\mathcal{X}}_P\in\mathbb{R}^{N\times L}$ denote the Discrete Four-\\ier Transform (DFT) of $\bm{X}_P$, in which each row of $\bm{\mathcal{X}}_P$ is the $L$-point DFT of the corresponding row of $\bm{X}_P$. Accordingly, the flipped spectrum satisfies, for $0\leq k\leq L-1$,
\begin{equation}
\bm{\mathcal{X}}'_P[k] = \bm{\mathcal{X}}_P[L-k].
\end{equation}
We apply the inverse DFT (IDFT) to the sum of $\bm{\mathcal{X}}'_P$ and $\bm{\mathcal{X}}_P$, obtaining the energy-equalized time series $\bm{X}^{\text{eq}}_{P}\in\mathbb{R}^{N\times L}$, i.e.,
\begin{equation}\label{samm}
\bm{X}^{\text{eq}}_{P}=\text{IDFT}(\bm{\mathcal{X}}_P+\bm{\mathcal{X}}'_P).
\end{equation}
As a result, the energy of the original low-energy components are increased to be commensurate with that of the high-energy\\ components. The above energy equalization process is conduc-\\ive to drawing the model’s attention for learning the previously neglected low-energy components.

One can see from (\ref{samm}) that there will be two separate energy peaks at both ends of the spectrum of $\bm{X}^{\text{eq}}_{P}$. We subsequently adopt Seasonal-Trend Decomposition widely used in previous work \cite{Zhou2022Fedformer,Wu2021Autoformer,Zeng2023Are} to decompose $\bm{X}^{\text{eq}}_{P}$ into the seasonal and trend components as
\begin{equation}
\bm{X}^{\text{trend}}_{P},\;\bm{X}^{\text{season}}_{P}=\text{Decom}(\bm{X}^{\text{eq}}_{P}),
\end{equation}
which correspond to these two energy peaks. $\bm{X}^{\text{trend}}_{P}$ represents the slow, non-periodic modifications in fMRI time series, e.g., scanner drift or gradual physiological shift. In contrast, $\bm{X}^{\text{season}}_{P}$ delineates the periodic or cyclical patterns, reflecting experim-\\ental task repetitions or intrinsic physiological rhythms. Then, by applying two FFNs to the two peaks separately and fusing their outputs, we obtain $\bm{X}^{\text{ts}}_{P}\in\mathbb{R}^{N\times D}$ as
\begin{equation}
\bm{X}^{\text{ts}}_{P} = \text{FFN}_{\text{trend}}(\bm{X}^{\text{trend}}_{P})+\text{FFN}_{\text{season}}(\bm{X}^{\text{season}}_{P}).
\end{equation}

We finally adopt a linear transformation to adjust the column dimension of $\bm{\mathcal{X}}'_P$ to $D$ (i.e., the column dimension of $\bm{H}^{\text{spat}}$ in (\ref{sss})), followed by an inverse operation of spectrum flipping to derive the temporal representation $\bm{H}^{\text{temp}}\in\mathbb{R}^{N\times D}$ (namely, the energy restoration process returns the energy to the original level), i.e., 
\begin{equation}
\begin{split}
\bm{\hat{\mathcal{X}}}'_P(n,:) &= \bm{\mathcal{X}}'_P(n,:) \bm{W}_P + \bm{b}_P,\\
\bm{\mathcal{X}}^{\text{ts}}_{P}&=\text{DFT}(\bm{X}^{\text{ts}}_{P}),\\
\bm{H}^{\text{temp}}&=\text{IDFT}(\bm{\mathcal{X}}^{\text{ts}}_{P}-\bm{\hat{\mathcal{X}}}'_P),
\end{split}
\end{equation}
where $\bm{\hat{\mathcal{X}}}'_P(n,:)$ and $\bm{\mathcal{X}}'_P(n,:)$ denote the $n$-th rows of $\bm{\hat{\mathcal{X}}}'_P$ and $\bm{\mathcal{X}}'_P$, respectively, $\bm{W}_P\in\mathbb{C}^{L\times D}$ is a complex-valued weight matrix, and $\bm{b}_P\in\mathbb{C}^{D}$ is a complex-valued bias vector.

\subsection{Spatio-Temporal Pattern Alignment}
The spatial and temporal representations ($\bm{H}^{\text{spat}}$ and $\bm{H}^{\text{temp}}$) exhibit substantial differences in the high-level semantic repre-\\sentation space. Directly concatenating them can lead to redun-\\dancy and inconsistency. Therefore, based on vector similarity retrieval, we design an SPA module to align $\bm{H}^{\text{spat}}$ with $\bm{H}^{\text{temp}}$, resulting in a global spatio-temporal representation, $\bm{H}^{\text{global}}\in \mathbb{R}^{N \times D}$, as illustrated in Fig. \ref{fig1}(d).

By three linear layers $\psi_q, \psi_v$, and $\psi_k$, we first project $\bm{H}^{\text{spat}}$ and $\bm{H}^{\text{temp}}$ to $\psi_q(\bm{H}^{\text{spat}}), \psi_k(\bm{H}^{\text{temp}})$, and $\psi_v(\bm{H}^{\text{temp}})$. Then, we compute the attention score matrix $\bm{M}$ as
\begin{equation}\label{sdadasd}
    \bm{M} = \text{Softmax}\left( \psi_q (\bm{H}^{\text{spat}}) (\psi_k (\bm{H}^{\text{temp}}))^\top \right),
\end{equation}
and perform channel-wise feature aggregation by restoring ch-\\annel dimension with the matrix multiplication of $\psi_v(\bm{H}^{\text{temp}})$ and $\bm{M}$. Finally, the output $\bm{H}^{\text{global}}$ is obtained as
\begin{equation}
    \bm{H}^{\text{global}} = \omega \left( \psi_v (\bm{H}^{\text{temp}}) \bm{M} \right) + \bm{H}^{\text{spat}},
\end{equation}
where $\omega$ stands for a linear layer. Via SPA, $\bm{H}^{\text{global}}$ explicitly captures both spatial and temporal dependencies among fMRI time series, enhancing the model's forecasting performance.

\subsection{Forecasting Head and Loss Function}
After extracting the latent representation $ \bm{H}^{\text{global}}$, we employ a prediction head to transform the feature space into the target fMRI time series $\bm{X}_F\in\mathbb{R}^{N\times T}$, i.e., for $1\leq n\leq N$,
\begin{equation}\label{prr}
    \hat{\bm{X}}_F(n,:) = \bm{H}^{\text{global}}(n,:)\bm{W}_F + \bm{b}_F,
\end{equation}
where $\bm{b}_F\in\mathbb{R}^{T}$ is a bias vector, and $\bm{W}_F \in\mathbb{R}^{D\times T}$ is a weight matrix. The model's parameters are optimized by minimizing the discrepancies between the predicted data $\hat{\bm{X}}_F$ in (\ref{prr}) and the ground-truth future data $\bm{X}_F$. For this purpose, we utilize the mean squared error (MSE) as the loss function; that is,
\begin{equation} \label{eq:mse_loss}
    \mathcal{L}_{\text{MSE}} = \frac{1}{N} \sum_{n=1}^{N} \left\| \hat{\bm{X}}_F(n,:)  - \bm{X}_F(n,:)  \right\|_2^2.
\end{equation}

\begin{table}[h]
\centering
\caption{Run duration and the number of frames (i.e., time points) per run for each fMRI paradigm.}
\label{txiao}
\renewcommand{\arraystretch}{1.25}
\resizebox{0.48\textwidth}{!}{
\begin{tabular}{lcc}
\hline
\textbf{fMRI paradigm}        & \textbf{Frames per run} & \textbf{Run duration (min:sec)} \\ \hline
\textbf{Resting-state}        & 1200                    & 14:33                           \\
\textbf{Working memory}       & 405                     & 5:01                            \\
\textbf{Emotional processing} & 176                     & 2:16                            \\
\textbf{Language processing}  & 316                     & 3:57                            \\
\textbf{Relational reasoning} & 232                     & 2:56                            \\
\textbf{Social cognition}     & 274                     & 3:27                            \\
\textbf{Gambling}             & 253                     & 3:12                            \\
\textbf{Motor execution}      & 284                     & 3:34                            \\ \hline
\end{tabular}}
\end{table}

\section{Experimental Results}\label{sec4}

\begin{table*}[t]
\centering
\caption{Performance comparison of different models on HCP-rs-fMRI and HCP-t-fMRI datasets. The mean is reported with the standard deviation shown in parentheses below. The best results are highlighted in bold.}
\label{tab:1}
\renewcommand{\arraystretch}{1.25}
\setlength{\tabcolsep}{5pt}
\resizebox{0.9\textwidth}{!}{
\begin{tabular}{ccccccccccc}
\toprule
\multirow[c]{2}{*}{\textbf{Dataset}} & \multirow[c]{2}{*}{\textbf{Metric}} &
\multicolumn{9}{c}{\textbf{Model}} \\
\cmidrule(lr){3-11}
& & Ours & iTransformer & DLinear & Amplifier & PatchTST & CrossFormer & FourierGNN & LSTM & RNN \\
\midrule
\multirow[c]{4}{*}{\textbf{HCP-rs-fMRI}\vspace{-1.5cm}} 
& MSE $\downarrow$ &  \makecell{\textbf{0.380} \\(0.004)} & \makecell{0.398\\ (0.002)} & \makecell{0.529 \\(0.012)} & \makecell{0.407 \\(0.006)} & \makecell{0.394\\ (0.003)} & \makecell{0.417\\ (0.001)} & \makecell{0.562\\ (0.005)}  & \makecell{0.602\\ (0.013)} & \makecell{0.592\\ (0.006)} \\ \addlinespace[4pt]
& MAE $\downarrow$ & \makecell{\textbf{0.461}\\ (0.004)} & \makecell{0.478\\ (0.002)} & \makecell{0.562\\ (0.009)} & \makecell{0.481\\ (0.005)} & \makecell{0.473\\ (0.004)} & \makecell{0.488\\ (0.003)} & \makecell{0.580\\ (0.004)} & \makecell{0.601\\ (0.011)} & \makecell{0.597\\ (0.008)}\\ \addlinespace[4pt]
& R $\uparrow$       & \makecell{\textbf{0.665}\\ (0.002)} & \makecell{0.631\\ (0.001)} & \makecell{0.446\\ (0.013)} & \makecell{0.627\\ (0.006)} & \makecell{0.635\\ (0.003)} & \makecell{0.628\\ (0.004)} & \makecell{0.432\\ (0.002)} & \makecell{0.355\\ (0.008)} & \makecell{0.379\\ (0.005)}\\ \addlinespace[4pt]
& R$^2 \uparrow$     & \makecell{\textbf{0.438}\\ (0.003)} & \makecell{0.397\\ (0.003)} & \makecell{0.199\\ (0.011)} & \makecell{0.390 \\(0.003)} & \makecell{0.402\\ (0.005)} & \makecell{0.396\\ (0.006)} & \makecell{0.187\\ (0.003)} & \makecell{0.120\\ (0.007)} & \makecell{0.143\\ (0.007)} \\
\addlinespace[2pt]
\midrule
\multirow{4}{*}{\textbf{HCP-t-fMRI}\vspace{-1.5cm}} 
& MSE $\downarrow$ & \makecell{\textbf{0.435}\\ (0.004)} & \makecell{0.488\\ (0.004)} & \makecell{0.633\\ (0.005)} & \makecell{0.460\\ (0.009)} & \makecell{0.485\\ (0.006)} & \makecell{0.480\\ (0.010)} & \makecell{0.705\\  (0.007)} & \makecell{0.681\\ (0.005)} & \makecell{0.663\\ (0.004)} \\ \addlinespace[4pt]
& MAE $\downarrow$ & \makecell{\textbf{0.484}\\ (0.006)} & \makecell{0.501\\ (0.002)} & \makecell{0.587\\ (0.007)} & \makecell{0.514\\ (0.015)} & \makecell{0.501\\ (0.003)} & \makecell{0.496\\ (0.008)} & \makecell{0.621\\  (0.005)} & \makecell{0.609\\ (0.010)} & \makecell{0.602\\ (0.008)} \\ \addlinespace[4pt]
& R $\uparrow$       & \makecell{\textbf{0.595}\\ (0.003)} & \makecell{0.583\\ (0.005)} & \makecell{0.376\\ (0.004)} & \makecell{0.582\\ (0.021)} & \makecell{0.586\\ (0.005)} & \makecell{0.589\\ (0.017)} & \makecell{0.208\\ (0.009)} & \makecell{0.292\\ (0.007)} & \makecell{0.320\\ (0.008)} \\ \addlinespace[4pt]
& R$^2 \uparrow$     & \makecell{\textbf{0.351}\\ (0.004)} & \makecell{0.338\\ (0.002)} & \makecell{0.142\\ (0.003)} & \makecell{0.333\\ (0.018)} & \makecell{0.341\\ (0.008)} & \makecell{0.347\\ (0.013)} & \makecell{0.043\\ (0.004)} & \makecell{0.076\\ (0.002)} & \makecell{0.101\\ (0.004)} \\
\bottomrule
\end{tabular}}
\end{table*}

\subsection{Datasets and Preprocessing}
In this study, we used the data from the Human Connectome Project (HCP) S1200 Release \cite{b18}, which included behavioral and multi-modal neuroimaging datasets collected from $1206$ healthy young adults aged $22$ to $35$ years. To ensure data integ-\\rity, a total of $862$ subjects ($409$ males and $453$ females) who had all four 3T MRI modalities (i.e., structural MRI, diffusion MRI, resting-state fMRI, and task fMRI) were selected. Deta-\\iled descriptions of their demographic characteristics and MRI acquisition parameters of the selected subjects can be readily found in the HCP protocol.

The resting-state fMRI data was acquired across two sessio-\\ns (R1 and R2) on separate days, and we solely used the fMRI data at the first session (R1) in our study. The task fMRI data\\ was acquired across seven task paradigms, including working memory, emotional processing, language processing, relational reasoning, social cognition, gambling, and motor execution. Two runs of the resting-state fMRI (R1) and each task fMRI were implemented, one with a left-to-right and the other with a right-to-left phase encoding. To avoid the potential influence of the phase encoding direction on our study, we used the left-to-right phase encoding runs of all fMRI data. In Table \ref{txiao}, we presented run duration and the number of time points (namely frames) per run for each fMRI paradigm.

All functional images underwent the HCP minimal preprocessing pipeline, including gradient distortion correction, head\\ motion correction, image distortion correction, spatial alignment to standard Montreal Neurological Institute (MNI) space, and intensity normalization. Moreover, further denoising procedures were performed, which involved regressing out motion parameters (and their first-order derivatives), removal of linear trends, and temporal band-pass filtering (0.01–0.1 Hz). After-\\wards, we parcellated all the preprocessed fMRI data into $268$ (i.e., $N=268$) ROIs based on the Shen functional atlas \cite{shenatlas}.

For simplicity, we denoted the above preprocessed resting-state fMRI data as HCP-rs-fMRI. We concatenated the preprocessed fMRI data from all the seven tasks along the temporal axis, and denoted the newly formed task fMRI data as HCP-t-fMRI. In this study, both HCP-rs-fMRI and HCP-t-fMRI were employed to evaluate the effectiveness of the proposed model for fMRI time series prediction, respectively.

\begin{figure*}[!t]
    \centering

    \begin{subfigure}{\textwidth}
        \centering
        \includegraphics[width=0.8\textwidth]{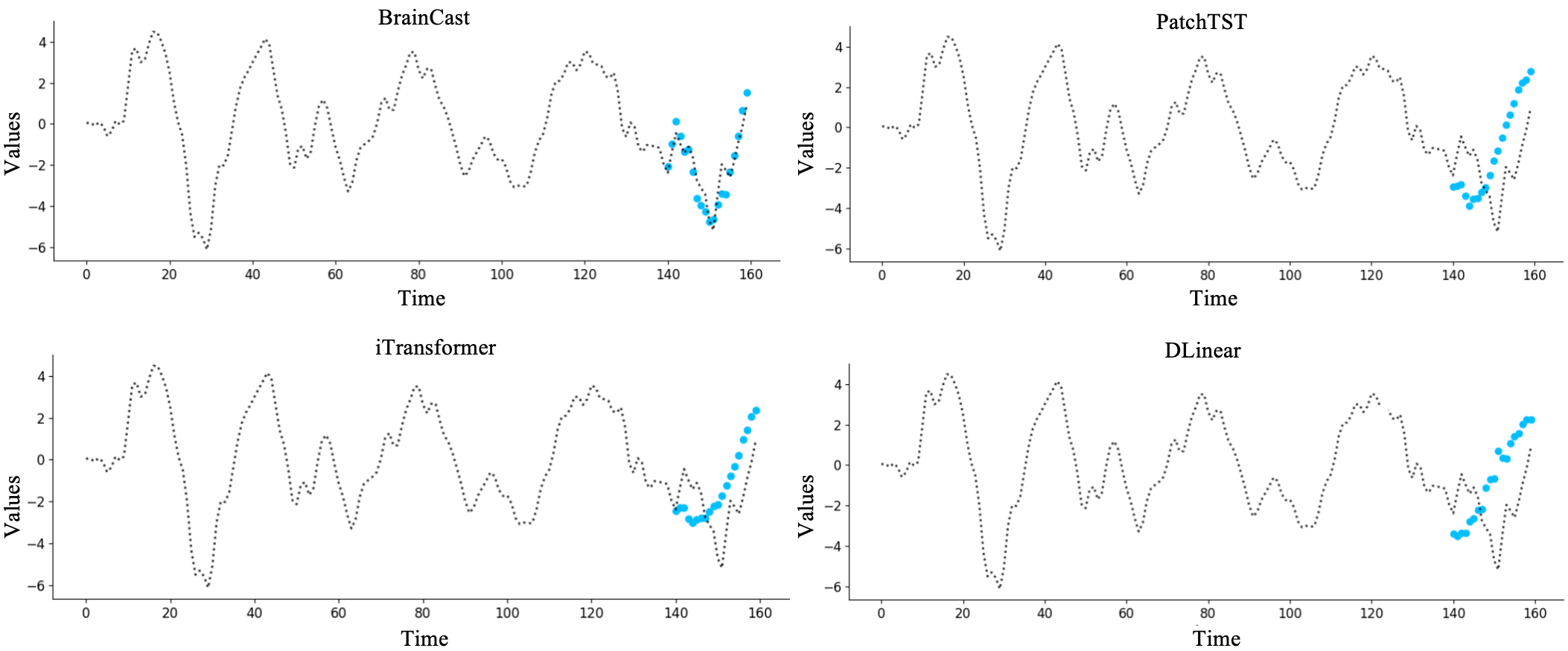}
        \caption{HCP-rs-fMRI}
    \end{subfigure}

    \vspace{0.4cm}

    \begin{subfigure}{\textwidth}
        \centering
        \includegraphics[width=0.8\textwidth]{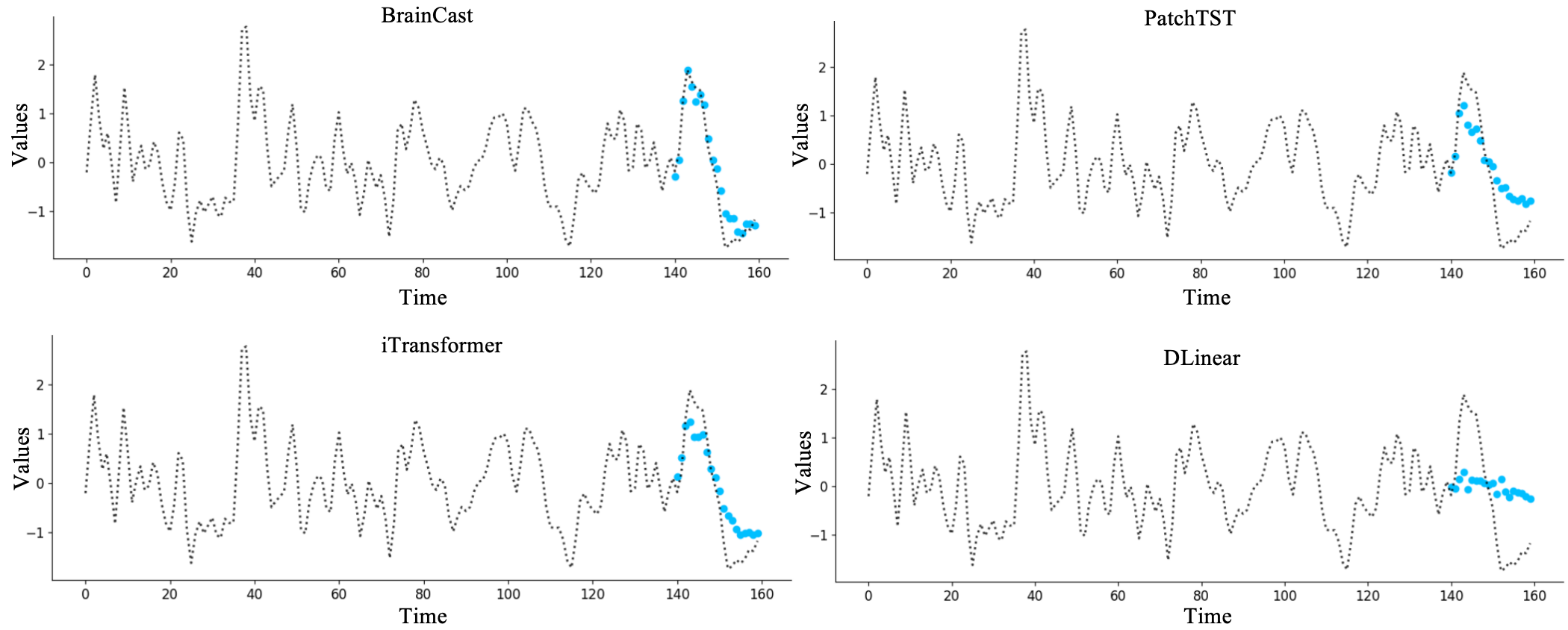}
        \caption{HCP-t-fMRI}
    \end{subfigure}

    \caption{Visualization of forecasting results on a randomly selected ROI of one test sample for HCP-rs-fMRI (a) and HCP-t-fMRI (b), respectively. The black dots denote the previous and ground-truth future data, and the blue dots denote the predicted data.}
    \label{vis_prediction}
\end{figure*}

\subsection{Implementation Details}
To avoid data leakage, we randomly split the whole subjects into $80\%$ for training, $10\%$ for validation, and $10\%$ for testing, whose generative samples by sliding window were used as the training, validation, and test sets, respectively. This was repea-\\ted five times to reduce the influence of data-splitting bias, and the mean and standard deviation (std) of the prediction results in all test sets were reported on the four widely used evaluation metrics, including mean absolute error (MAE), mean squared error (MSE), Pearson's correlation coefficient (R), and coeffic-\\ient of determination (R²).

For parameter optimization, we selected the RMSprop optimizer, which adaptively adjusted the learning rate based on a moving average of squared gradients. The initial learning rate was set to $1\times 10^{-5}$, and training was performed with a batch size of 64. To mitigate overfitting and improve generalization, an early stopping strategy was implemented, by which training was halted when the validation loss did not improve for $5$ con-\\secutive epochs. The main hyperparameters in our model were given as follows. The lengths of the previous and future fMRI time series over a sliding window were $L=140$ and $T=20$, respectively; and the stride was $s=20$. Our model contained $2$ SIAformer layers (namely $G=2$) and $8$ multi-head attention heads, and the embedding dimension was set to $D=512$. Be-\\fore feeding the time series data into our model, we normalized the data to stabilize training and improve robustness. 

\subsection{Forecasting Performance}
With the use of HCP-rs-fMRI and HCP-t-fMRI, respective-\\ly, we compared the fMRI time series forecasting performance of our model (i.e., BrainCast) and $8$ highly regarded baselines, including cutting-edge Transformer models (such as PatchTST \cite{b22}, CrossFormer \cite{Zhang2023Crossformer}, and iTransformer \cite{Liu2023itransformer}), efficient linear neural network models (such as DLinear \cite{Zeng2023Are} and Amplifier \cite{Fei2025Amplifier}), classic recurrent neural network models (such as LSTM \cite{b9} and RNN \cite{b10}), and graph neural network models (such as FourierGNN \cite{b24}).

As exhibited in Table \ref{tab:1}, BrainCast consistently achieved the best forecasting performance on these two datasets in terms of both the error-based metrics (MSE and MAE) and correlation-based metrics (R and R²). For example, on HCP-rs-fMRI, our model reached superior performance across all the metrics, de-\\creasing MSE by $\geq 0.014$ and MAE by $\geq 0.012$, and increas-\\ing R by $\geq 0.030$ and R² by $\geq 0.036$. Similarly, on HCP-t-fMRI, our model outperformed the the second-best models by an obvious margin. Moreover, one can see that the Transformer models (also including ours) achieved better performance than\\ the others. In particular, such recurrent neural network architectures, i.e., LSTM and RNN, struggled to capture long-range dependencies in time series forecasting, leading to inferior per-\\formance. Although FourierGNN incorporated spectral repres-\\entations of time series, its performance is limited in the fMRI time series forecasting tasks. This may be attributed to the co-\\mplex spatio-temporal dynamics of brain neural activities that cannot be satisfactorily captured by frequency-domain analysis alone. By contrast, our model can practically characterize not only spatial interactions between ROIs but temporal variations within every ROI, highlighting its potential in modeling neural dynamics in multi-region fMRI time series.

To further demonstrate the superiority of our model, we vis-\\ualized the forecasting results of our model, PatchTST, iTransformer, and DLinear in Fig. \ref{vis_prediction}. For easy visualization, we pres-\\ented the forecasting result of a randomly selected ROI of one test sample with respect to HCP-rs-fMRI and HCP-t-fMRI, re-\\spectively. As exhibited in Fig. \ref{vis_prediction}, our model provided the most accurate predictions on future fMRI time series, substantially outperforming the other models, where the predicted trajectories not only followed the overall trend of the ground truth but also aligned well with local fluctuations, including rising and falling patterns. In addition, we observed that despite the pre-\\sence of rapid transitions or high-frequency oscillations in the time series, our model maintained stable forecasting accuracy. It indicates that our model is capable of learning both low-fre-\\quency trends and high-frequency variations in neural activity, which are essential to accurately modeling brain dynamics.

\subsection{Ablation Studies}
\begin{figure}[htbp]
    \centering
    \begin{subfigure}{\columnwidth}
        \centering
        \includegraphics[width=0.9\columnwidth]{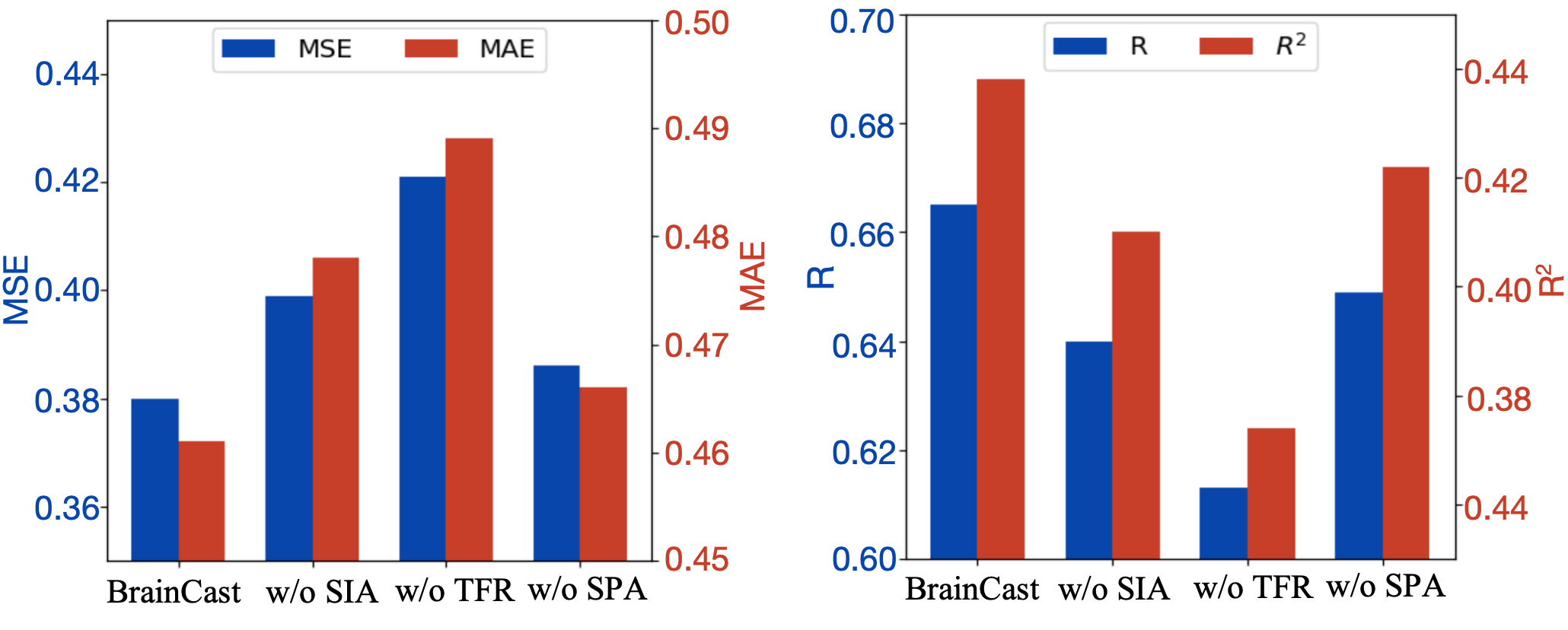}
        \caption{HCP-rs-fMRI}
    \end{subfigure}
    
    \vspace{0.4cm} 
           
    \begin{subfigure}{\columnwidth}
        \centering
        \includegraphics[width=0.9\columnwidth]{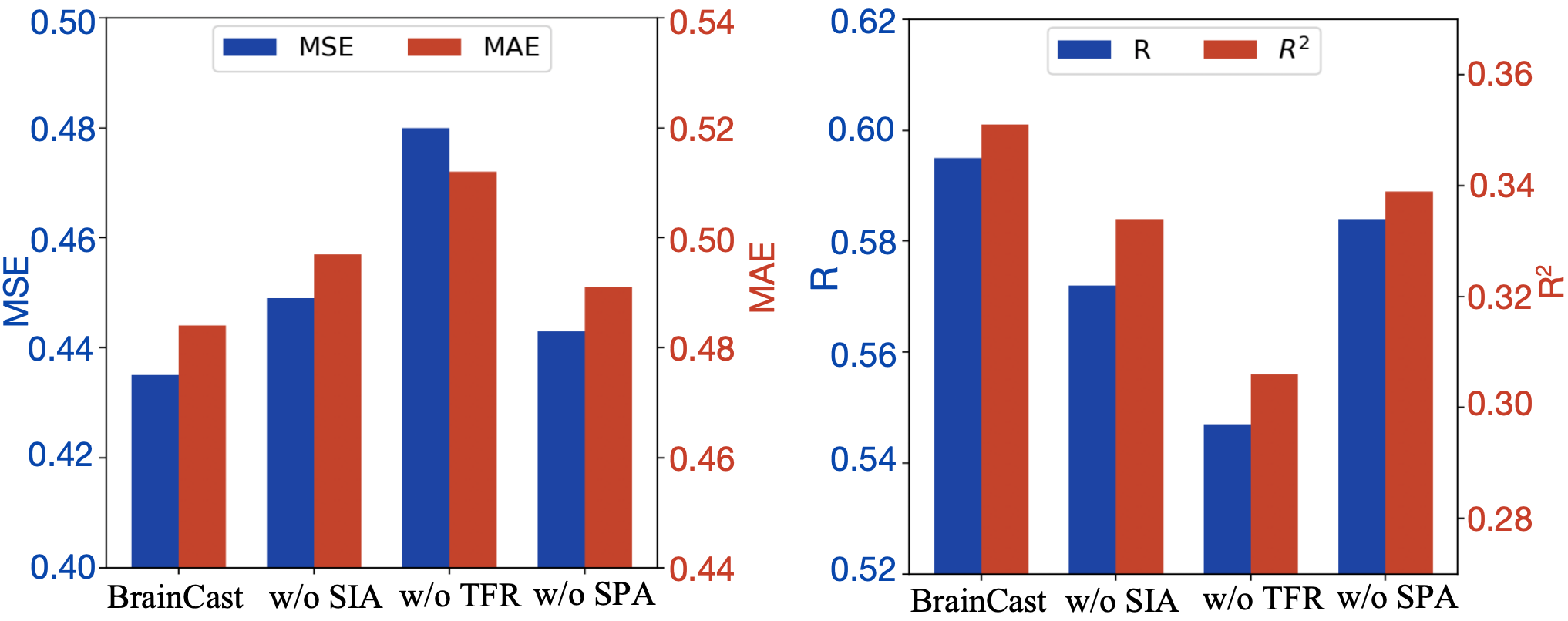}
        \caption{HCP-t-fMRI}
    \end{subfigure}
    \caption{Ablation results for the SIA, TFR, and SPA modules in our BrainCast on both HCP-rs-fMRI (a) and HCP-t-fMRI (b), respectively. The dark blue bars correspond to the left y-axis, whereas the dark red bars correspond to the right y-axis.}
    \label{fig:ablation}
\end{figure}

To verify whether the SIA, TFR, and SPA modules designed in our model were essential, we conducted ablation experime-\\nts on both HCP-rs-fMRI and HCP-t-fMRI, as shown in Fig. \ref{fig:ablation}. One can see that removing any module consistently caused forecasting performance degradation across all evaluation metrics on both datasets. Specifically, removing TFR resulted in the most inferior performance, with MSE and MAE increasing while R and R² decreasing dramatically. This demonstrates the importance of modeling temporal dynamics for accurate fMRI time series forecasting. In addition, removing SIA also brought about significant performance decline, highlighting the necess-\\ity of capturing long-range spatial interactions between ROIs. In comparison, SPA exerted a relatively small yet stable impact on performance, substantiating its vital role in aligning spatio-temporal representations output from SIA and TFR. Therefore, all such modules contributed to BrainCast in a complementary manner, and the full model achieved the best performance.

\begin{figure*}[!t]
\centerline{\includegraphics[width=0.75\textwidth]{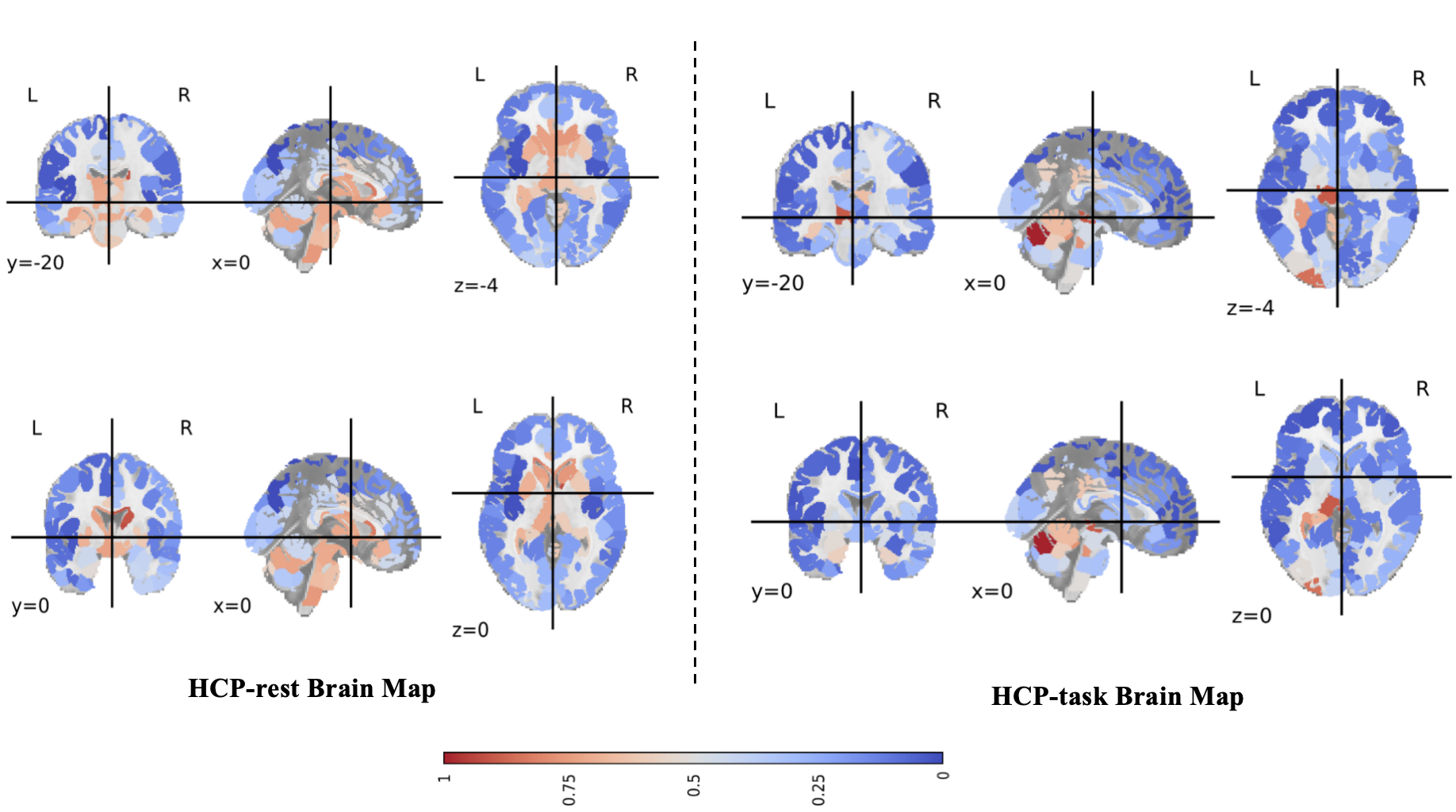}}
\caption{Visualization of the cortical attention maps in the coronal, sagittal, and axial planes, where red indicates higher attention scores (i.e., greater contribution to the model’s prediction) and blue indicates lower contributions.}
\label{fig5}
\end{figure*}

\section{Discussions} \label{sec5}

\subsection{Visualization of Attention Maps}
To elucidate the neurobiological underpinnings of our model for fMRI time series forecasting with HCP-rs-fMRI and HCP-t-fMRI, we visualized the cortical attention maps (i.e., the atte-\\ntion score matrices in (\ref{sdadasd})), respectively, by mapping the atte-\\ntion scores to the corresponding ROIs on the cortical surface; see Fig. \ref{fig5}. Notably, the attention map from HCP-rs-fMRI reve-\\aled pronounced left–right hemispheric symmetry, with consp-\\icuous emphasis on homotopic regions, while that from HCP-t-fMRI lacked such symmetry and displayed a heterogeneous distribution of activated areas.

\begin{figure}[htbp]
    \centering
    \begin{subfigure}[t]{0.45\columnwidth}
        \centering
        \includegraphics[width=\linewidth]{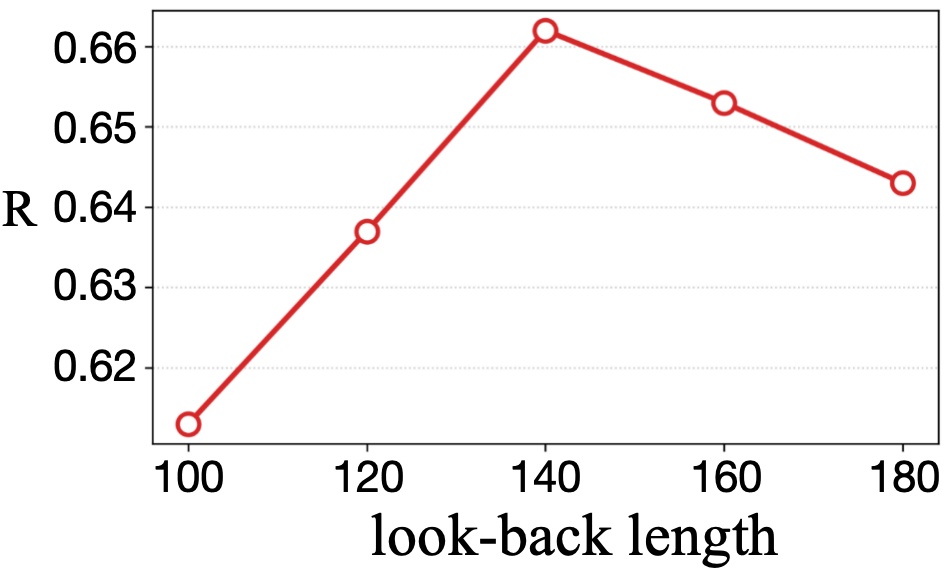}
        \caption{HCP-rs-fMRI}
        \label{sensitive_rest}
    \end{subfigure}
    \hfill
    \begin{subfigure}[t]{0.45\columnwidth}
        \centering
        \includegraphics[width=\linewidth]{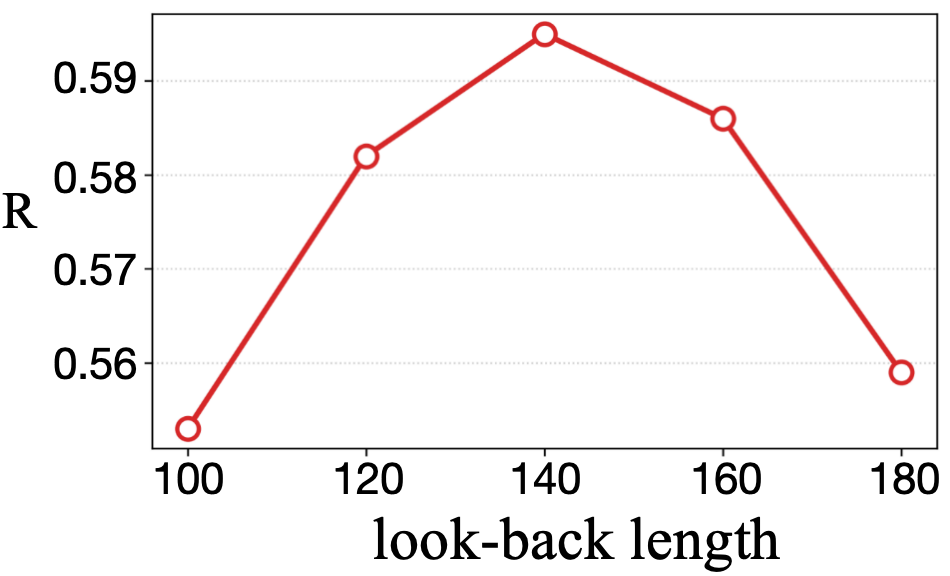}
        \caption{HCP-t-fMRI}
        \label{sensitive_task}
    \end{subfigure}

    \caption{The R results of BrainCast with varying look-back len-\\gths (i.e., the length of the previous fMRI time series $L$) and the fixed future length (i.e., the length of the future fMRI time series $T=20$) on both HCP-rs-fMRI and HCP-t-fMRI.}
    \label{sensitive}
\end{figure}

This striking bilateral symmetry during rest aligns well with prior findings on homotopic functional connectivity: the voxel-\\mirrored homotopic connectivity metric provides a quantitative measure of robust interhemispheric synchronization in regions, including cingulate, precentral, and postcentral gyri, therefore characterizing the intrinsic functional integration of homotopic areas at rest \cite{b45}. By contrast, task-driven states may attenuate this symmetric pattern of reliance. Task engagement typically recruits network-specific and lateralized circuits tailored to sti-\\mulus demands, consequently reducing the prominence of ho-\\motopic correspondence in predictive contributions. Actually, dynamic reconfiguration of resting-state networks—especially the default mode and dorsal attention subnetworks—has been demonstrated, with rest promoting bilateral coherence whereas attention states elicit more lateralized, task-specific connectivity patterns \cite{b46}.

Together, these observations suggest that during rest, homo-\\topic connectivity provides a stable, symmetric scaffold explo-\\ited by the model for prediction, potentially reflecting under-\\lying interhemispheric pathways. Under naturalistic stimulation, however, brain dynamics appear to become asymmetric, context-sensitive, attenuating the relative contribution of mirror regions and favoring specialized processing.

\subsection{Influence of Look-Back Length}





To investigate the impact of the length of the previous fMRI time series (i.e., $L$, also called look-back length) in BrainCast on the forecasting performance, we conducted additional exp-\\eriments with varying $L\in\left\{100, 120, 140, 160, 180\right\}$ and fixed future length $T=20$ (i.e., the length of the future fMRI time series) on both HCP-rs-fMRI and HCP-t-fMRI. As shown in Fig. \ref{sensitive}, the fMRI time series forecasting performance measured\\ by R initially improves with increasing $L$, achieves a peak at an intermediate value of $L=140$, and then gradually declines as $L$ continues to increase. This trend suggests a trade-off bet-\\ween contextual sufficiency and information redundancy. Incr-\\easing the look-back length initially enhances temporal model-\\ing by providing richer historical context; however, when the\\ look-back length becomes excessively long, the forecasting pe-\\rformance degrades as the model struggles to effectively capt-\\ure increasingly complex temporal variations in extended seq-\\uences, and informative patterns might be diluted by redundant information. Conversely, overly short look-back lengths afford insufficient contextual information, limiting the model’s ability to characterize temporal dependencies and thereby impairing forecasting accuracy.

\subsection{Application in Establishing Brain-Cognition Relationships}
To explore the potential application of fMRI time series fo-\\recasting and evaluate the added value with BrainCast, we implemented designed experiments to investigate brain–cognition relationships. Specifically, we used two types of data (i.e., the previous fMRI time series and the prolonged fMRI time series\\ extended by BrainCast) to predict the Penn Matrix Reasoning Test (PMAT) scores. As part of the HCP \cite{b18}, the PMAT was carried out in the computerized neurocognitive battery (CNB) to quantify nonverbal reasoning-related cognitive ability \cite{b50}. In this setting, the previous fMRI time series of length $140$ were augmented to the corresponding prolonged ones of length $160$, with the last $20$-point fMRI time series being forecasted\\ by BrainCast. For the PMAT score prediction, we compared the root mean squared error (RMSE) derived from the previous and prolonged fMRI time series across a broad range of regres-\\sion approaches, including convolutional neural networks (e.g., BrainNetCNN \cite{b27}) , graph convolutional networks (e.g., GCN \cite{b31}), MLP \cite{b48}, and support vector regression (SVR) \cite{b49}, as\\ reported in Table \ref{tab:2}. One can readily see that compared with the previous fMRI time series, the prolonged fMRI time series consistently improved the cognitive prediction performance ac-\\ross all the regression approaches. This finding underscores the\\ potential applicability and impact of fMRI time series foreca-\\sting on neuroscience and clinical research.

\begin{table}[!h]
\centering
\caption{PMAT prediction performance on HCP-rs-fMRI, measured in terms of RMSE ($\text{mean} \pm \text{standard deviation}$).}
\label{tab:2}
\renewcommand{\arraystretch}{1.25}
\resizebox{0.48\textwidth}{!}{
\begin{tabular}{lcc}
\toprule
\textbf{Method} & \textbf{HCP-rs-fMRI (previous)} & \textbf{HCP-rs-fMRI (prolonged)} \\
\midrule
BrainNetCNN & 5.0155 $\pm$ 0.0017 & 4.8925 $\pm$ 0.0303 \\
MLP         & 4.8778 $\pm$ 0.0382 & 4.7443 $\pm$ 0.0221 \\
GCN         & 4.8473 $\pm$ 0.0150 & 4.7383 $\pm$ 0.0470 \\
SVR         & 4.7698 $\pm$ 0.0121 & 4.6361 $\pm$ 0.0598 \\
\bottomrule
\end{tabular}}
\end{table}

\section{Conclusion} \label{sec6}
In this paper, we proposed BrainCast, a new spatio-temporal\\ alignment driven framework for whole-brain fMRI time series forecasting. Through integrating the SIA, TFR, and SPA mod-\\ules, our BrainCast can jointly capture inter-ROI spatial depen-\\dencies and intra-ROI temporal dynamics in a complementary and principled manner. Extensive experiments on both HCP-rs\\-fMRI and HCP-t-fMRI demonstrated that
BrainCast consiste-\\ntly outperformed state-of-the-art time series forecasting models across several performance evaluation metrics. In addition, by augmenting the previous fMRI time series using BrainCast, we observed consistent improvements in downstream cognitive ability prediction. These results highlight the potential applica-\\bility and added value of fMRI time series forecasting, which contributes to enhancing brain–cognition analyses, particularly in settings with limited scan durations. Future work will focus on subject-specific variability, multi-modal data fusion, and the clinical validation of BrainCast in neurological applications.

\bibliographystyle{IEEEtran}
\bibliography{refs}

\end{document}